\documentclass[11pt]{article}
\usepackage{coling2018}
\usepackage{times}
\usepackage{latexsym}
\usepackage{booktabs}
\usepackage{todonotes}
\usepackage{linguex}
\usepackage{tikz-dependency}
\usepackage{url}

\usepackage[usenames,dvipsnames]{pstricks}
\usepackage{epsfig}
\usepackage{enumitem} 
\usepackage[hang,flushmargin,symbol]{footmisc} 

\makeatletter

\def\citealt{\def\citename##1{{\frenchspacing##1} }\@internalcitec}

\def\@citexc[#1]#2{\if@filesw\immediate\write\@auxout{\string\citation{#2}}\fi
  \def\@citea{}\@citealt{\@for\@citeb:=#2\do
    {\@citea\def\@citea{;\penalty\@m\ }\@ifundefined
       {b@\@citeb}{{\bf ?}\@warning
       {Citation `\@citeb' on page \thepage \space undefined}}%
{\csname b@\@citeb\endcsname}}}{#1}}

\def\@internalcitec{\@ifnextchar [{\@tempswatrue\@citexc}{\@tempswafalse\@citexc[]}}

\def\@citealt#1#2{{#1\if@tempswa, #2\fi}}

\makeatother

\title{All Roads Lead to UD:
Converting Stanford and Penn Parses to English Universal Dependencies with Multilayer Annotations}

\author{Siyao Peng \\
Department of Linguistics \\
  Georgetown University \\
  {\tt sp1184@georgetown.edu} \\\And
  Amir Zeldes \\
Department of Linguistics \\
  Georgetown University \\
  {\tt amir.zeldes@georgetown.edu} \\}
\date{}

\begin{document}
\maketitle
\begin{abstract}
We describe and evaluate different approaches to the conversion of gold standard corpus data from Stanford Typed Dependencies (SD) and Penn-style constituent trees to the latest English Universal Dependencies representation (UD 2.2). Our results indicate that pure SD to UD conversion is highly accurate across multiple genres, resulting in around 1.5\% errors, but can be improved further to fewer than 0.5\% errors given access to annotations beyond the pure syntax tree, such as entity types and coreference resolution, which are necessary for correct generation of several UD relations. We show that constituent-based conversion using CoreNLP (with automatic NER) performs substantially worse in all genres, including when using gold constituent trees, primarily due to underspecification of phrasal grammatical functions.\blfootnote{This work is licensed under a Creative Commons Attribution 4.0 International License. License
details: http://creativecommons.org/licenses/by/4.0/. This paper is a cross-submission and has already been accepted for presentation at LAW-MWE-CxG-2018}
\end{abstract}

\section{Introduction}
In the past two years, the Universal Dependencies project (UD, \citealt{NivreEtAl2017}), offering freely available dependency treebanks with a unified annotation scheme in over 50 languages, has grown rapidly, allowing for cross-linguistic comparison and computational linguistics applications. At the same time, because of its rapid growth and the need to negotiate annotation schemes across languages, annotating large resources from scratch in the latest UD standard is challenging, not only because of the annotation effort, but also because guidelines may change mid-way, and data and annotator training must be revisited to match the latest developments. Instead, a large number of projects within UD capitalize on existing treebanks converted from constituent treebanks (in English usually using CoreNLP, \citealt{ManningEtAl2014}) or other dependency schemes, meaning that for those projects that are not annotated directly in UD, changes to the UD guidelines generally mean adapting an existing converter framework.

In this paper, we concentrate on English dependency treebanking, which has been dominated by data converted from Penn Treebank-style constituent trees (cf. \citealt{BiesFergusonKatzEtAl1995}). We compare results of constituent treebank conversions with results from converting English dependency data annotated using the older (and by now frozen) Stanford Typed Depenendencies (hence SD, \citealt{MarneffeManning2013}). Specifically, we will be working with the freely available Georgetown University Multilayer corpus (GUM, \url{http://corpling.uis.georgetown.edu/gum/}), which we have converted to the latest UD standard (as of UD version 2.2). The paper has several goals:

\begin{enumerate}
\item To describe and evaluate the accuracy of gold standard SD to UD conversion (SD2UD)
\item To explore the necessary layers of annotation for generating gold UD from gold SD data, including information that is not strictly present in the syntactic parse
\item Comparing conversions from SD source data and constituent tree source data
\item Making a substantial new English resource, with over 85,000 tokens in 8 genres, available in UD	
\end{enumerate}

We will show that while rule-based SD to UD conversion is already highly accurate, it must also rely on multiple annotation layers outside of the parse proper if the full range of dependencies is targeted. For the third goal in particular, our evaluation of the converted UD product reveals that `native' dependency data in English differs from converted constituents in several ways, including the presence of some rare labels and the proportion of non-projective dependencies.

\section{Corpora}

The main corpus used in this paper is the Georgetown University Multilayer corpus (GUM, \citealt{Zeldes2017}), a freely available corpus covering data from eight English genres: news, interviews, how-to guides, travel guides, academic writing, biographies, fiction and web forum discussions. The corpus is annotated by students at Georgetown University\footnote{For an analysis of annotation quality and genre differences within the corpus, see \newcite{ZeldesSimonson2016}} and currently contains 101 documents, with over 85,000 tokens, annotated for:

\begin{itemize}
\item Multiple POS tags (Penn tags, \citealt{Santorini1990}, TreeTagger tags and CLAWS5 tags, \citealt{GarsideSmith1997}), as well as lemmatization
\item Sentence segmentation and rough speech act (based on SPAAC, \citealt{LeechEtAl2003}) 
\item Document structure (paragraphs, headings, etc.), ISO date/time annotations and speaker information
\item Gold SD dependencies and automatic constituent parses based on gold POS tags
\item Information status (given, accessible and new, based on \citealt{DipperGoetzeSkopeteas2007})
\item Entity and coreference annotation, including bridging anaphora
\item Discourse parses in Rhetorical Structure Theory (\citealt{MannThompson1988})
\end{itemize}

A second English corpus we will be comparing this data to in Section \ref{other_corp} is the English Web Treebank (\citealt{BiesMottWarnerEtAl2012}, \citealt{SilveiraEtAl2014}), containing over 1,170 documents with over 250,000 tokens in five genres: blog posts, e-mails, newsgroup discussions, online answer forums and online reviews. This corpus was originally annotated using Penn-style constituent trees and converted to UD using CoreNLP (\citealt{SchusterManning2016}), with subsequent scripts and manual corrections producing the version now available in UD V2.2. 

\section{Method}

In this section we focus on describing our approach to converting SD parses to UD with and without supplemental information from further layers of annotation. The evaluation in Section \ref{eval} will compare these scenarios with several conversion scenarios from constituent trees.

\subsection{SD conversion rules}

Our conversion process comprises three parts: 

\begin{enumerate}
\item a preprocessing step pulling in information from annotation layers outside of the syntax tree proper
\item the main rule-based conversion
\item a postprocessing step in which punctuation is attached using the freely available udapi API (\citealt{PopelZabokrtskyVojtek2017})
\end{enumerate}

This section concentrates on the main, syntactic rule-based conversion, while the next section focuses on information brought in from other annotation layers. 

The main step uses a configurable rule-based converter called DepEdit\footnote{Available at \url{https://corpling.uis.georgetown.edu/depedit/} and via PyPI (\texttt{pip install depedit}).} which allows the definition of conversion rules, each having three components: 1. a set of key-value pairs denoting regular expressions matching targeted token properties; 2. a set of relations which must hold between these tokens; and 3. instructions on how to alter token properties when the rule is matched. Some example rules are given in Table \ref{depedit}.

\begin{table*}
\begin{center}
\begin{tabular}{|l|l|l|}
\hline \bf attributes & \bf relations & \bf actions \\ 
\hline

func=/dobj/ & none & \#1:func=obj  \\
func=/.*/;func=/\^{}cc\$/;func=/\^{}conj\$/ & \#1$>$\#2;\#1$>$\#3 & \#3$>$\#2 \\
func=/prep/;pos=/\^{}W.*/;func=/pcomp/ & \#1$>$\#3;\#3$>$\#2 & \#2:func=pobj;\#1$>$\#2;\#2$>$\#3;\#3:func=rcmod\\


\hline

\end{tabular}
\end{center}
\caption{\label{depedit} Examples of DepEdit rules}
\end{table*}

The first example illustrates a trivial renaming rule, in which the SD label \textit{dobj} is renamed to UD \textit{obj}: the definition in the first column matches any token with a function label matching \texttt{/dobj/}, no relations are imposed (\texttt{none}), and the action specifies that the first (and only) token in the definition, \#1, should have its function label set to \textit{obj}. Similar rules are used to create Universal POS tags, which is almost trivial, since the corpus already contains gold Penn Treebank-style POS tags and lemmas. However, in some cases, dependency relations must be consulted too, e.g. the verb `be' must be given the AUX tag as a copula or auxiliary, and otherwise VERB; determiners (e.g. \textit{that}) become DET when modifying nouns, but are PRON when used independently; etc.

\begin{figure}[h]
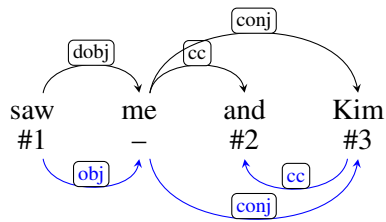

\centering
\begin{dependency}[arc edge, arc angle=80, text only label, label style={above}]
\begin{deptext}[column sep=.7cm]
saw \& me \& and \& Kim \\
\#1 \& -- \& \#2 \& \#3 \\
\end{deptext}
\depedge{1}{2}{dobj}
\depedge{2}{3}{cc}
\depedge[edge below, edge style={blue}, label style={text=blue}]{4}{3}{cc}
\depedge{2}{4}{conj}
\depedge[edge below, edge style={blue}, label style={text=blue}]{1}{2}{obj}
\depedge[edge below, edge style={blue}, label style={text=blue}]{2}{4}{conj}
\end{dependency}
\caption{Converting coordination from SD to UD}\label{conj_ex}
\end{figure}

The second example in Table \ref{depedit} is more complex and changes the graph in Figure \ref{conj_ex} from the coordinating conjunction `and' being governed by the first conjunct (SD guidelines) to being governed by the second (UD V2.2 guidelines). The attribute definitions first specify `any function' (\texttt{func=/.*/}), then for a second token (separated by `;') that its function must be \textit{cc} (coordinating conjunction), followed by a third token labeled \textit{conj}. The relations column then specifies that token \#1 governs \#2 and that it also governs \#3. Finally the actions column specifies that \#3 should now govern \#2, leaving unchanged the fact that \#1 governs \#3. The process of applying these two rules is shown for a fragment in Figure \ref{conj_ex}, where the source (SD) graph is rendered above the tokens, and the result (UD) below, rendered in blue.

The third rule handles free relative clauses, and targets WH pronouns governed by a preposition and \textit{pcomp}, in constructions such as ``an expectation of$_{\#1}$ what$_{\#2}$ to do$_{\#3}$'', which should be converted to a relative clause (in SD, \textit{rcmod}). Note that since this rule occurs before conversion of prepositions to the UD label \textit{case} and relatives to \textit{acl}, SD labels are still used in this rule. POS substitutions are also cascaded, meaning rules can initially refer to Penn tags, and later on to UPOS tags.\footnote{Morphological features, by contrast, are generated at the end of the process using CoreNLP, as is the case for EWT. Their accuracy is not evaluated in this paper.}

The most current set of conversion rules, numbering nearly 100 items, can be found along with conversion utilities is freely available online.\footnote{\url{https://github.com/amir-zeldes/gum/blob/dev/_build/utils/stan2uni.ini}}

\subsection{Using multilayer annotations}\label{multilayer}

The availability of several kinds of non-syntactic gold annotations in GUM allows us to refine the conversion process further. While it could be argued that syntax trees should not contain non-syntactic information to begin with, UD parses do in fact integrate information which seems to be not completely syntactic, and more so than SD: specifically, as we will see below, factors such as ontological entity types, coreference information and presence of errors or disfluencies all affect the analysis in UD. This can be viewed as an advantage of UD trees once the information is available, but also as an unfair requirement from parsers and converters attempting to generate data in the UD scheme.

One of the most widespread changes not recognizable from pure SD dependencies is the conversion of SD \textit{nn} (noun modified noun) into one of two structures: \textit{compound} for nominal compounds with internal syntactic structure and \textit{flat} for headless multi-word expressions that are not part of the closed list receiving the label \textit{fixed}. In practice, the \textit{flat} label in English usually translates to proper nouns supplying names. 

\begin{figure}[h]
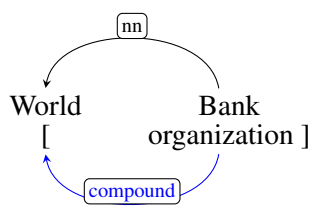

\centering
\begin{dependency}[arc edge, arc angle=80, text only label, label style={above}]
\begin{deptext}[column sep=.7cm]
World \& Bank \\
\lbrack \& organization \rbrack \\
\end{deptext}
\depedge{2}{1}{nn}
\depedge[edge below, edge style={blue}, label style={text=blue}]{2}{1}{compound}
\end{dependency}
\caption{\label{world_bank}Converting `World Bank' (organization)}
\end{figure}

\begin{figure}[h]
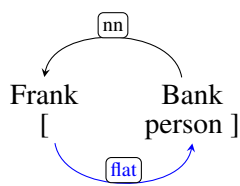

\centering
\begin{dependency}[arc edge, arc angle=80, text only label, label style={above}]
\begin{deptext}[column sep=.7cm]
Frank \& Bank \\
\lbrack \& person \rbrack \\
\end{deptext}
\depedge{2}{1}{nn}
\depedge[edge below, edge style={blue}, label style={text=blue}]{1}{2}{flat}
\end{dependency}
\caption{\label{frank_bank}Converting `Frank Bank' (person)}
\end{figure}

The large majority of \textit{flat} cases correspond to names of persons, while most named non-persons retain a syntactic head (usually on the right).\footnote{The other main category containing \textit{flat} names is place names, but the majority of multi-word place names are nevertheless headed, and therefore labeled \textit{compound}. A discussion on whether or not proper names such as `Kim King' should be treated as non-headed, or arbitrarily annotated as head-initial, is beyond the scope of this paper.} This means that knowing entity types can be crucial. For example, knowing that \textit{World Bank} is an \textit{organization} in Figure \ref{world_bank} induces the \textit{compound} relation between the two tokens; by contrast, in Figure \ref{frank_bank}, \textit{Frank Bank}, annotated as a \textit{person} entity on another annotation layer results in a \textit{flat} UD annotation.\footnote{An anonymous reviewer has remarked that the difference between Frank Bank and World Bank is arguably only a convention. This is certainly a valid point in general, but there is also some reason to consider differences between the structures, as codified in UD: while \textit{World Bank} is without a doubt a kind of `bank', the decision whether \textit{Frank Bank} is a kind of `Frank' or `Bank' is more arbitrary. This becomes more crucial when nested compounds are considered, since multi-part names can be seen as truly flat, but compounds like \textit{World Bank Federation} are recursive and right-headed.} The preprocessing step reads entity annotation information from parallel files and flags the (SD) head of each entity mention with its entity type, which is then used in the DepEdit conversion rules. The entity's head token is matched by finding a token in the entity span which is either the sentence root, or is governed by a non-punctuation parent from outside the span.

\begin{figure}[h]
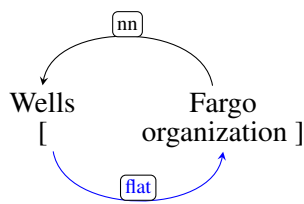

\centering
\begin{dependency}[arc edge, arc angle=80, text only label, label style={above}]
\begin{deptext}[column sep=.7cm]
Wells \& Fargo \\
\lbrack  \& organization \rbrack \\
\end{deptext}
\depedge{2}{1}{nn}
\depedge[edge below, edge style={blue}, label style={ text=blue}]{1}{2}{flat}
\end{dependency}
\caption{\label{wells_fargo}Analysis of `Wells Fargo' (organization)}
\end{figure}

However, the mapping between dependencies and entity types is not one-to-one, meaning some errors are inevitable even with gold entity information. For example, some company names arguably do not exhibit internal syntactic structure and should be annotated as \textit{flat} in UD, for example \textit{Wells Fargo} in Figure \ref{wells_fargo}. Currently, our automatic conversion will erroneously label such cases as \textit{compound} (see Section \ref{err_analysis} for error analysis).

A second type of information is required by the introduction of the label \textit{dislocated} in UD. Although dislocation, shown in \ref{disloc}, is ostensibly a syntactic operation, it appears very much like any kind of topicalization, as shown in \ref{tpc}. Both types are annotated as \textit{dep} in the GUM SD annotations, for lack of a better label.

\ex. We like pets. $[$\textbf{My neighbors}$_{dislocated}]$, their pets drive $[$\textbf{them}$_{obj}]$ nuts\label{disloc}

\ex. We like $[$\textbf{canned foods}$]$. My neighbors$_{dep}$, their pets eat $[$\textbf{them}$_{obj}]$ every day\label{tpc}

The semantic criterion distinguishing these two examples is that the dislocated node must be coreferential with a dependent of the verb (`them'=`neighbors').\footnote{One anonymous reviewer has suggested that \textit{dislocated} should be used for all fronted dependents, even if they are not realized a second time, citing a Japanese example from the UD guidelines. While we believe that marking fronting in general is interesting, and could perhaps be done using sublabels (e.g. \textit{obj:front}), we feel that marking fronted English arguments, as in ``him, I like'' with \textit{dislocated} is counter-intuitive, since it makes a verb such as `like' appear to be missing an object. The practice in other English corpora, and specifically in EWT, has been to only mark \textit{dislocated} in the presence of a second realization of the argument. The difference in the practice for Japanese may be due to the fact that in that language a second mention as a pronoun is usually omitted, and the closest equivalent of such a pronoun is therefore a zero-mention.} Because GUM has gold coreference annotations available, the preprocessing step again introduces a feature into the SD data which indicates a coreference ID for each coreferent nominal head, and nodes with the same coreference ID and syntactic head are changed from \textit{dep} to \textit{dislocated}.

\begin{figure}[h]
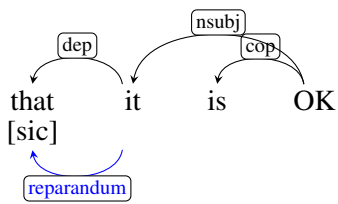

\centering
\begin{dependency}[arc edge, arc angle=80, text only label, label style={above}]
\begin{deptext}[column sep=.7cm]
that  \& it \&  is \&  OK \\
\lbrack sic\rbrack \&  \& \&  \\
\end{deptext}
\depedge{2}{1}{dep}
\depedge[edge below, edge style={blue}, label style={text=blue, below}]{2}{1}{reparandum}
\depedge{4}{3}{cop}
\depedge{4}{2}{nsubj}
\end{dependency}
\caption{\label{repair}labeling errors as \textit{reparandum}}
\end{figure}

Another type of information that may be seen as not purely syntactic is the presence of disfluencies. Though rare in written data, UD reserves a label for repairs in disfluencies or false starts, which can be used for both spoken and written data. The guidelines apply the label \textit{reparandum} to the head of the `aborted' part of the sentence, which is attached to the repair. The SD annotations in GUM follow the same structure, but apply the default label \textit{dep}, meaning that the presence of the disfluency needs to be detected. This is accomplished in the preprocessing step by checking GUM's TEI XML annotations that denote all types of errors in the corpus with \textit{$<$sic$>$} tags. Although these tags do not indicate the nature of the error or the repair, any occurrences of the \textit{dep} label inside an error and governed from outside of it are converted into \textit{reparandum}, as shown for the false start in Figure \ref{repair}.

We also use the \textit{$<$sic$>$} annotations to create a feature in the MISC column defined by the CoNLL-U format with the value \texttt{Typo=Yes}, as used to denote errors in other UD treebanks. These are not necessarily always cases of repair, but also cases of unusual or non-standard grammatical constructions or even orthographic anomalies such as non-matching quotation marks, as in \ref{mets}.

\ex. so quite a few fans \textit{$<$sic$>$}\textbf{known}\textit{$<$/sic$>$} about the ``Mets Poet\textit{$<$sic$>$}\textbf{'}\textit{$<$/sic$>$}\label{mets}

The same MISC column is also used to indicate whether tokens are followed by spaces using the feature \texttt{SpaceAfter=No}. The latter feature is also derived from the TEI annotations, where the presence of the tag \textit{$<$w$>$} indicates multiple tokens spelled together as one orthographic word.

\section{Evaluation}\label{eval}

\subsection{Experimental setup}

We compare UD conversions from SD and constituent annotations in several scenarios on a total of 8,300 tokens, comprising just over 1,000 tokens from each genre in GUM, or about 10\% of the corpus, for which we created manually checked gold UD parses. To evaluate constituent to UD (`C2UD') conversion accuracy, we created three constituent parsed versions of the same data using the Stanford Parser: one based on gold-tokenized plain text, one from data with gold POS tags, and the third, also parsed from gold POS tags, but then manually corrected for errors. The manually corrected constituent parses do not introduce empty categories such as PRO or traces, but do use function labels that may be critical for conversion, such as S-TPC (for fronted direct speech, common e.g. in fiction) and NP-VOC, NP-TMP and NP-ADV for vocative, temporal and other adverbial NPs.\footnote{The data used for the evaluation, including different versions of constituent parses, is available at \url{https://github.com/gucorpling/GUM_UD_LAW2018}.} C2UD conversion was carried out using CoreNLP 3.9.1, which uses built-in NER and heuristic time expression recognition, but is not completely up-to-date with the current UD standard. We therefore apply trivial renaming of labels where needed and two heuristic corrections: all coordinating conjunctions (labeled \textit{cc}) are attached to the original target of the \textit{conj} relation, so that they point right to left; and all nominal modifiers of verbs (labeled \textit{nmod}) are re-labeled as \textit{obl}.

In scoring correct conversion we focus on two metrics: attachment accuracy ignoring punctuation tokens (since punctuation is automatically attached using udapi, \citealt{PopelZabokrtskyVojtek2017}, and errors are therefore by-products of other attachment errors), and label accuracy, including punctuation (since some punctuation symbols are occasionally used for non-punctuation functions). Because there are some differences in the label subtypes produced by CoreNLP and GUM (e.g. \textit{obl:tmod}, \textit{nmod:npmod}), we ignore subtypes for the evaluation and focus on main label types.

\subsection{Results}

\begin{figure*}[hbt]
\centering
\fbox{
\includegraphics[width=\textwidth]{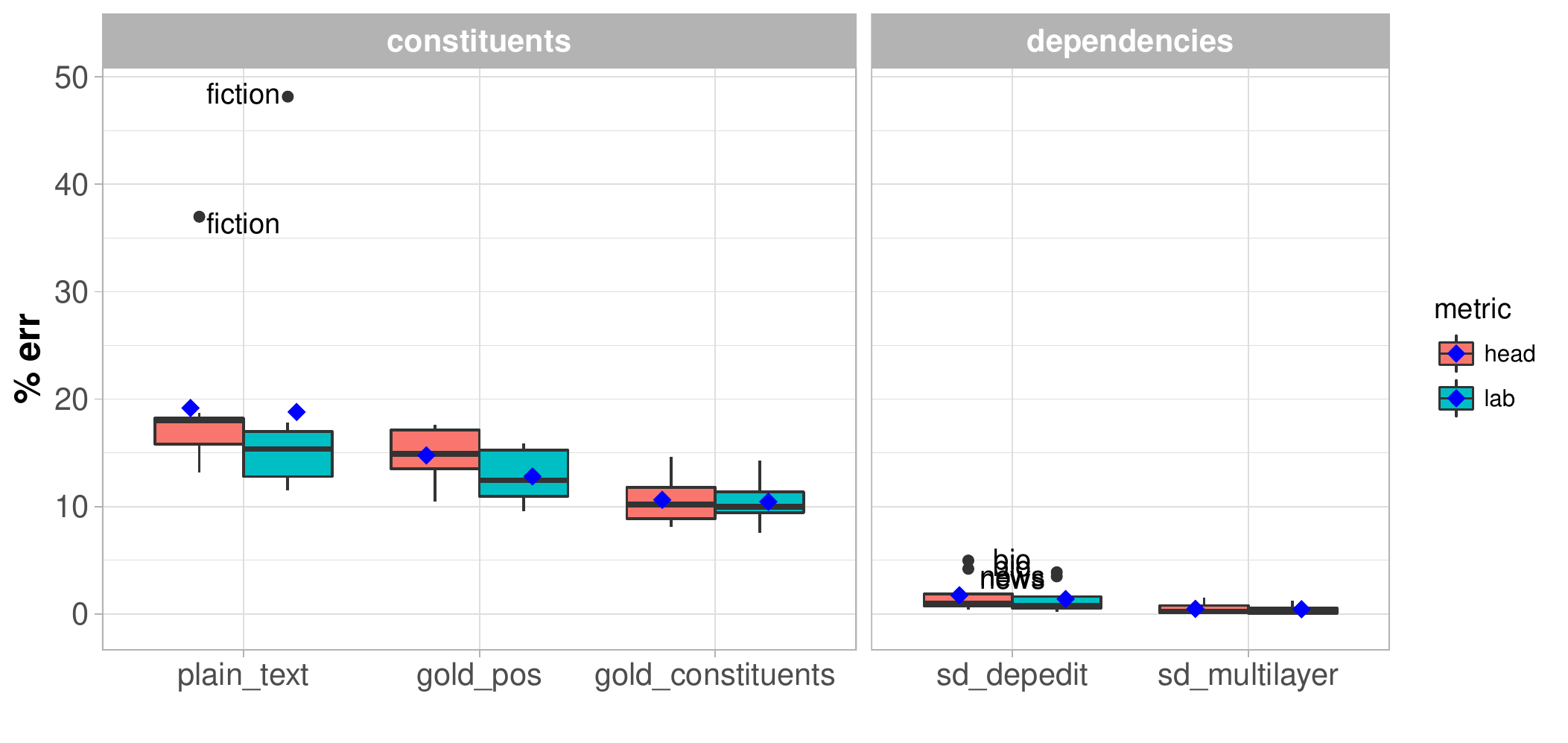}}
\caption{Error rates}
\label{err_rates}
\end{figure*}

Figure \ref{err_rates} shows boxplots for the range of error rates across documents from different genres in five scenarios (tokenwise micro-averaged global means are given in blue diamonds), each splits  into two metrics: head and label accuracy.

In the best scenario, converting SD to UD with parallel multilayer information, conversion errors are very few, at 0.45\%/0.42\% of tokens (head/label errors). When multilayer annotations are removed, accuracy suffers somewhat, but is still rather good, with under 1.73\%/1.38\% errors. The more difficult genres for pure SD conversion are news and biographies, though only by a little: since these genres contain many multi-token proper names, correct conversion relies more on entity types, which cannot be recognized in the pure DepEdit conversion, but are available to the multilayer conversion.

Comparing SD with constituent conversions, error rates become more substantial. Errors in the `plain text' scenario are just under 20\%; keeping in mind that the Stanford parser is trained on Wall Street Journal data, this is in line with previous results on parsing accuracy for out-of-domain constituent to dependency conversion \cite{ChoiPalmer2010}.

The not much better results for gold POS and gold constituents, by contrast, may seem surprising initially, since in general, constituents do identify the main argument structure relations, such as subjects and objects. However, a range of decisions cannot be made deterministically  without semantic knowledge. Some of these might be avoided more reliably in datasets containing empty categories (traces, pro-forms) and more category sub-labels (e.g. PP-CLR, etc. see \citealt{BiesFergusonKatzEtAl1995}), but the GUM constituents, even in their cleanest form, are based on CoreNLP constituent parses, which do not contain these. 

Outliers in the `plain text' scenario correspond to fiction texts, which frequently contained different Unicode quotation marks that are mistagged by CoreNLP. Gold POS tags remove the issue, as the `gold pos' scenario shows. Nevertheless, even with gold constituents, mean error rates remain at 10.62\%/10.44\%. To understand the limitations of both constituent and SD to UD conversions, we examine some specific error patterns in the next section. 

\subsection{Error analysis}\label{err_analysis}

To understand what conversion errors need to be avoided, we first consider the difficulties in C2UD conversion. Table \ref{err_labs} shows the top 3 most frequent gold labels causing attachment and labeling errors for gold constituents, pure SD, and multilayer SD conversion. 

\begin{table}[hbt]
\begin{center}
\begin{tabular}{|l|r|l|r|l|}

\hline \bf scenario  & \multicolumn{2}{c|}{\textbf{head errs}} & \multicolumn{2}{c|}{\textbf{lab errs}}\\
\hline
C2UD  & 84 & nsubj & 130 & obl\\
(gold)  & 82 & nmod & 74 & nmod\\
 &  71 & conj & 62 & conj\\
 \hline
SD & 37 & flat & 37 & flat\\
(pure)  & 10 & nmod & 8 & obl\\
  & 8 & appos & 7 & nsubj\\
  \hline
SD & 8 & compound & 9 & compound\\
(multi) & 6 & nmod & 7 & obl\\
 & 6 & flat & 6 & nmod\\

\hline

\end{tabular}
\end{center}
\caption{\label{err_labs} Top 3 gold labels showing head and label errors in three scenarios}
\end{table}

In C2UD, even given gold constituents, many pure phrase labels are highly ambiguous with respect to their exact function. This is especially true for fronted NPs without function labels, which can be fronted arguments (\textit{dislocated}, \textit{obj}, \textit{iobj}), a spatio-temporal adverbial (\textit{advmod:npmod}), a vocative (\textit{vocative}) and more. These are sometimes misidentified as subjects, leading to true gold subjects being misrecognized (objects are not as susceptible due to their position inside VPs). Conversely, the label \textit{obl} is most often mislabeled, usually in cases where prepositional modifiers of nominal or adjectival predicates are not recognized and labeled \textit{nmod}. In general, whenever phrases are extraposed, their attachment site cannot be predicted accurately in the absence of trace annotations, and these are most often labeled \textit{nmod} and \textit{obl}.

In third place, coordination is the next most problematic construction, due to the fact that PTB brackets do not explicitly mark coordination (except for Unlike Coordinate Phrases, labeled UCP). As a result, some non-standard but frequent types of coordination are missed, such as using `/' for `or' (common in web data), `et al.' (common in academic data) and unmarked coordination or lists using commas, which can look like appositions in constituent trees. All of these distinctions are represented directly in SD, which is conceptually much closer to UD, and thus these errors are virtually absent in the SD scenarios. 

The errors in the Pure SD scenario are dominated by missing \textit{flat} relations in proper names, due to the lack of entity recognition; guessing that all SD \textit{nn} relations are UD \textit{compound} is the safer choice. The confusion of \textit{obl} and \textit{nmod} features here as well, but is much less frequent, due to gold attachment data in the SD parses which is usually trivial to convert to UD. Errors in appositions and subject relations are almost only by-products of incorrect name conversions, since the head token of the entire name is wrongly selected. In the Multilayer SD scenario, we see the over-generation errors in producing \textit{compound} relations for non-person names -- these are cases like `Well Fargo', which should in fact be \textit{flat} as well.

Additionally we note that the conversion from constituents is qualitatively missing some rare labels. These include cases that require the extra-syntactic knowledge described in section \ref{multilayer}, such as \textit{dislocated} and \textit{reparandum}, but also the label \textit{goeswith}, which indicates multiple tokens belonging to one `word' but spelled apart, and the \textit{vocative} label, which could hypothetically be guessed or derived directly if constituents include the NP-VOC subtype. While all of these labels represent rare phenomena, their exclusion from the constituent conversion output is problematic.

\begin{figure*}[hbt]
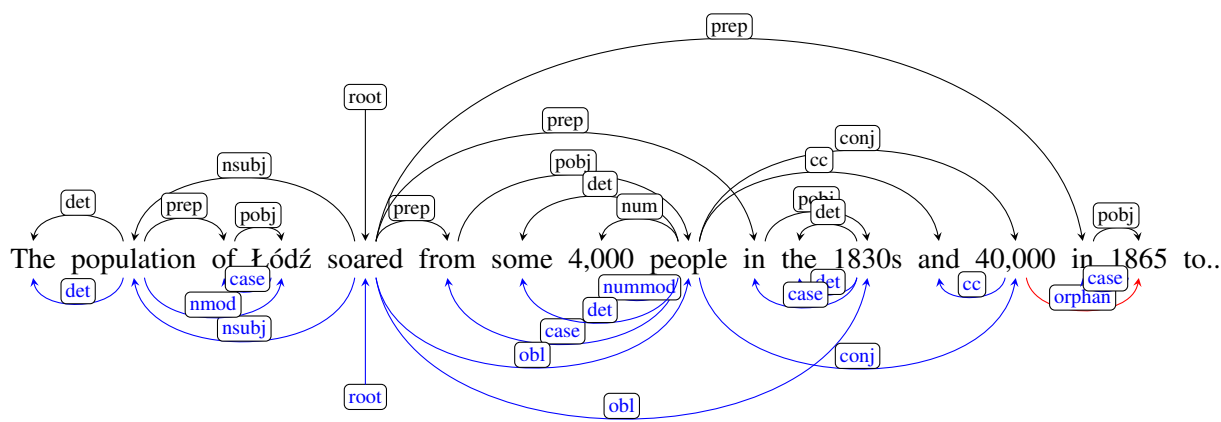

\centering
\begin{dependency}[arc edge, arc angle=80, text only label, label style={above}]
\begin{deptext}[column sep=.03cm]
The \& population \& of \& \L{}\'{o}d\'{z} \& soared \& from \& some \& 4,000 \& people \& in \& the \& 1830s \& and \& 40,000 \& in \& 1865 \& to..\\
\end{deptext}
\deproot{5}{root}
\depedge{2}{1}{det}
\depedge{5}{2}{nsubj}
\depedge{2}{3}{prep}
\depedge{3}{4}{pobj}
\depedge{5}{6}{prep}
\depedge{6}{9}{pobj}
\depedge{9}{7}{det}
\depedge{9}{8}{num}
\depedge{5}{10}{prep}
\depedge{10}{12}{pobj}
\depedge{12}{11}{det}
\depedge{9}{14}{conj}
\depedge{9}{13}{cc}
\depedge{5}{15}{prep}
\depedge{15}{16}{pobj}
\deproot[edge below, edge style={blue}, label style={text=blue}]{5}{root}
\depedge[edge below, edge style={blue}, label style={text=blue}]{5}{2}{nsubj}
\depedge[edge below, edge style={blue}, label style={text=blue}]{2}{1}{det}
\depedge[edge below, edge style={blue}, label style={text=blue}]{2}{4}{nmod}
\depedge[edge below, edge style={blue}, label style={text=blue}]{4}{3}{case}
\depedge[edge below, edge style={blue}, label style={text=blue}]{9}{6}{case}
\depedge[edge below, edge style={blue}, label style={text=blue}]{5}{9}{obl}
\depedge[edge below, edge style={blue}, label style={text=blue}]{9}{8}{nummod}
\depedge[edge below, edge style={blue}, label style={text=blue}]{9}{7}{det}
\depedge[edge below, edge style={blue}, label style={text=blue}]{12}{11}{det}
\depedge[edge below, edge style={blue}, label style={text=blue}]{5}{12}{obl}
\depedge[edge below, edge style={blue}, label style={text=blue}]{12}{10}{case}
\depedge[edge below, edge style={blue}, label style={text=blue}]{9}{14}{conj}
\depedge[edge below, edge style={blue}, label style={text=blue}]{14}{13}{cc}
\depedge[edge below, edge style={red}, label style={text=blue}]{14}{16}{orphan}
\depedge[edge below, edge style={blue}, label style={text=blue}]{16}{15}{case}

\end{dependency}
\caption{Example showing the \textit{orphan} relation, not represented in SD}\label{orphan}
\end{figure*}

Finally we wish to point out one label that is currently not generated by any of our scenarios: the label \textit{orphan}, which indicates promotion of a token to dominate the child of a missing coordinate parent. The construction, shown in Figure \ref{orphan}, is not directly expressible using SD relations and as such has been annotated somewhat unfaithfully by reference to the non-elliptical parent in the example.

In the SD original (black edges), the population of Lodz is said to have soared from 4,000 people in the 1830s (two prepositional modifiers of `soared'), and from 40,000 in 1865, to some other number. The inclusion of both the `1830s' and the `1865' as modifiers of `soared' makes it seem as if both years apply at the same time. UD adds the relation \textit{orphan} to express a second elliptical `soared', which would have connected `40,000' and `1865' (``and $[$soared from$]$ 40,000 in 1865...''). Though the UD solution seems clearly superior to the SD one, it is difficult to derive automatically without further annotations indicating the semantic structure, or using labels other than those found in SD.\footnote{One reviewer has suggested that a better analysis of Figure \ref{orphan} is to treat the two phrases after `from' as a coordination, making the years part of the same constituents as the numbers of people, i.e.: ``$[$from $[[$some 4000 people in the 1830s$]$ and $[$40000 in 1865$]]]$...''. However this solution incorrectly groups together the years and numbers, despite the fact that `4000 people in the 1830s' is not a constituent. Although the UD analysis with \textit{orphan} is imperfect in not explicitly duplicating the node corresponding to `soar', such an explicit analysis could be made using the optional Enhanced UD representation, which includes `copy nodes'.}

\subsection{Comparison with other corpora}\label{other_corp}

Aside from the accuracy of the conversion, we would like to suggest that there are some qualitative and quantitative differences between UD English data from `native dependencies' and `native constituents'. Qualitatively, some UD labels cannot be reliably produced via conversion, and are therefore absent in the initial C2UD result. This applies as noted above to the labels \textit{dislocated} and \textit{reparandum}, as well as \textit{vocative} if conversion from NP-VOC is not used, though these labels can be reintroduced manually, as has been done in subsequent corrections to the EWT, for example.

Quantitatively, we note that non-projective dependencies, which are generally rare in English, are more frequent in SD2UD conversion than in C2UD. Table \ref{non_proj} shows frequencies for non-projective dependencies, excluding punctuation cases, across the entire EWT and GUM corpora in two scenarios: first, automatic C2UD conversion with CoreNLP is compared for both corpora. Then the current, partially manually corrected UD EWT V2.2 is compared with the multilayer conversion from SD for GUM, and the proportion of non-projectivity in the original gold SD data is given for comparison.

\begin{table}[hbt]
\begin{center}
\begin{tabular}{|l|r|r|r|}
\hline
& \textbf{C2UD} & \textbf{UD V2.2 (corrected)} & \\
\hline
EWT & 0.34\% & 0.46\% & \\
\hline
& \textbf{C2UD} &  \textbf{UD V2.2 (from SD multi)} & \textbf{original SD} \\
\hline
GUM & 0.29\% & 0.79\% & 0.63\% \\
\hline

\end{tabular}
\end{center}
\caption{\label{non_proj} Non-projectivity in GUM and EWT.}
\end{table}

The table shows that C2UD conversion creates less non-projectivity than human corrected or SD converted data, which is perhaps unsurprising. A more surprising result is that the manually corrected EWT contains substantially less non-projectivity than the SD2UD version of GUM. This could be due to genre differences, though the difference is rather substantial (almost double). If the numbers in EWT in fact under-represent the actual non-projectivity in the data, then this may be an indication that the less projective nature of the `native constituents' EWT is shining through to the end result in the current UD version of the data. Finally we note that, at least for GUM, the conversion from gold SD to UD introduces further non-projectivity when compared to the original. A preliminary inspection of the constructions responsible for this suggests that coordinating conjunctions (the label \textit{cc}) pointing backwards in UD instead of forwards in SD is responsible for the largest increase in cases of non-projectivity, but further study is needed to understand the extent and distribution of non-projective constructions generated by each scheme.

\section{Discussion and outlook}

The approach taken in this paper confirms that SD annotations are conceptually quite close to UD, making a purely rule-based conversion highly accurate. At the same time, we have shown that for some less frequent labels, information from annotation layers beyond the pure syntax tree is needed, and this reduces error rates from around 1.5\% to closer to 0.4\%. By contrast, conversion from constituent trees, even when these are manually checked, still results in around 10\% errors (excluding punctuation).

An advantage of the present approach is the relative ease of the ability to change rules quickly as UD guidelines evolve: because the SD inventory is frozen, information that is derivable from the parse tree and further layers of annotation can be harnessed to produce the latest UD annotation scheme. It is also conceivable that retaining both SD and UD parses of the corpus can offer complementary information in some cases where UD collapses distinctions, e.g. between verbal modifiers labeled \textit{vmod} in SD and other adverbial clauses labeled \textit{advcl}. 

One of the main limitations of the SD scheme with respect to producing the current UD standard is the lack of a function corresponding to \textit{orphan}. This relation is also difficult for parsers to analyze correctly (see \citealt{SchusterNivreManning2018} for recent progress), meaning on the one hand that it is difficult to recognize automatically, and on the other, that it is desirable to include it in treebanks precisely in order to improve the availability of training data for such constructions.

In the future we would like to harness even more information from other layers in the corpus, both to enrich UD annotations with data in the MISC field and to validate annotation correctness. For example, using RST discourse parses available in GUM, we can draw on knowledge that certain clauses are \textit{purpose} clauses to distinguish controlled to-infinitives (\textit{xcomp}) from infinitival adverbial clauses (\textit{advcl}). We are currently considering which other annotations can be used to enrich and improve the quality of UD corpora for which other concurrent annotations are available.

\section*{Acknowledgments}

We would like to thank the reviewers, Nathan Schneider, and the UD community, for valuable comments on previous versions of this work, and the growing GUM annotation team for making their annotations available in the corpus -- this resource could not have been created without their contributions. For the latest list of GUM contributors, please see the corpus website at \url{http://corpling.uis.georgetown.edu/gum/}.

\bibliography{stan2ud.bib}

\begin{thebibliography}{}

\bibitem[\protect\citename{Bies \bgroup et al.\egroup
  }1995]{BiesFergusonKatzEtAl1995}
Ann Bies, Mark Ferguson, Karen Katz, and Robert MacIntyre.
\newblock 1995.
\newblock Bracketing guidelines for {Treebank II} style. {Penn Treebank
  Project}.
\newblock {CIS Technical Report MS-CIS-95-06}, University of Pennsylvania.

\bibitem[\protect\citename{Bies \bgroup et al.\egroup
  }2012]{BiesMottWarnerEtAl2012}
Ann Bies, Justin Mott, Colin Warner, and Seth Kulick.
\newblock 2012.
\newblock {E}nglish {W}eb {T}reebank.
\newblock ~ {LDC2012T13}, Linguistic Data Consortium, Philadelphia, PA.

\bibitem[\protect\citename{Choi and Palmer}2010]{ChoiPalmer2010}
Jinho~D. Choi and Martha Palmer.
\newblock 2010.
\newblock Robust constituent-to-dependency conversion for {E}nglish.
\newblock In {\em Proceedings of the 9th International Workshop on Treebanks
  and Linguistic Theories (TLT 2010)}, pages 55--66, Tartu, Estonia.

\bibitem[\protect\citename{de Marneffe and Manning}2013]{MarneffeManning2013}
Marie-Catherine de~Marneffe and Christopher~D. Manning.
\newblock 2013.
\newblock Stanford typed dependencies manual.
\newblock Technical report, Stanford University.

\bibitem[\protect\citename{Dipper \bgroup et al.\egroup
  }2007]{DipperGoetzeSkopeteas2007}
Stefanie Dipper, Michael G\"{o}tze, and Stavros Skopeteas.
\newblock 2007.
\newblock Information structure in cross-linguistic corpora: Annotation
  guidelines for phonology, morphology, syntax, semantics, and information
  structure.
\newblock {\em Interdisciplinary Studies on Information Structure}, 7.

\bibitem[\protect\citename{Garside and Smith}1997]{GarsideSmith1997}
Roger Garside and Nicholas Smith.
\newblock 1997.
\newblock A hybrid grammatical tagger: {CLAWS4}.
\newblock In Roger Garside, Geoffrey Leech, and Anthony McEnery, editors, {\em
  Corpus Annotation: Linguistic Information from Computer Text Corpora}, pages
  102--121. Longman, London.

\bibitem[\protect\citename{Leech \bgroup et al.\egroup }2003]{LeechEtAl2003}
Geoffrey Leech, Tony McEnery, and Martin Weisser.
\newblock 2003.
\newblock {SPAAC} speech-act annotation scheme.
\newblock Technical report, Lancaster University.

\bibitem[\protect\citename{Mann and Thompson}1988]{MannThompson1988}
William~C. Mann and Sandra~A. Thompson.
\newblock 1988.
\newblock Rhetorical {S}tructure {T}heory: Toward a functional theory of text
  organization.
\newblock {\em Text}, 8(3):243--281.

\bibitem[\protect\citename{Manning \bgroup et al.\egroup
  }2014]{ManningEtAl2014}
Christopher~D. Manning, Mihai Surdeanu, John Bauer, Jenny Finkel, Steven~J.
  Bethard, and Davide McClosky.
\newblock 2014.
\newblock The {S}tanford {CoreNLP} natural language processing toolkit.
\newblock In {\em Proceedings of ACL 2014: System Demonstrations}, pages
  55--60, Baltimore, MD.

\bibitem[\protect\citename{Nivre \bgroup et al.\egroup }2017]{NivreEtAl2017}
Joakim Nivre, \v{Z}eljko Agi\'{c}, Lars Ahrenberg, Maria~Jesus Aranzabe,
  Masayuki Asahara, Aitziber Atutxa, Miguel Ballesteros, John Bauer, Kepa
  Bengoetxea, Riyaz~Ahmad Bhat, Eckhard Bick, Cristina Bosco, Gosse Bouma, Sam
  Bowman, Marie Candito, G\"{u}l\c{s}en~Cebiro\u{g}lu Eryi\u{g}it, Giuseppe
  G.~A. Celano, Fabricio Chalub, Jinho Choi, \c{C}a\u{g}r{\i} \c{C}\"{o}ltekin,
  Miriam Connor, Elizabeth Davidson, Marie-Catherine de~Marneffe, Valeria
  de~Paiva, Arantza~Diaz de~Ilarraza, Kaja Dobrovoljc, Timothy Dozat, Kira
  Droganova, Puneet Dwivedi, Marhaba Eli, Toma\v{z} Erjavec, Rich\'{a}rd
  Farkas, Jennifer Foster, Cl\'{a}udia Freitas, Katar{\'{\i}}na
  Gajdo\v{s}ov\'{a}, Daniel Galbraith, Marcos Garcia, Filip Ginter, Iakes
  Goenaga, Koldo Gojenola, Memduh G\"{o}k{\i}rmak, Yoav Goldberg,
  Xavier~G\'{o}mez Guinovart, Berta~Gonz\'{a}les Saavedra, Matias Grioni,
  Normunds Gr{\=u}z{\={\i}}tis, Bruno Guillaume, Nizar Habash, Jan Haji\v{c},
  Linh~H{\`a} M{\~y}, Dag Haug, Barbora Hladk\'{a}, Petter Hohle, Radu Ion,
  Elena Irimia, Anders Johannsen, Fredrik J{\o}rgensen, H\"{u}ner
  Ka\c{s}{\i}kara, Hiroshi Kanayama, Jenna Kanerva, Natalia Kotsyba, Simon
  Krek, Veronika Laippala, L{\^e}~H\`{\^{o}}ng, Alessandro Lenci, Nikola
  Ljube\v{s}i\'{c}, Olga Lyashevskaya, Teresa Lynn, Aibek Makazhanov,
  Christopher Manning, C\u{a}t\u{a}lina M\u{a}r\u{a}nduc, David Mare\v{c}ek,
  H\'{e}ctor~Mart{\'{\i}}nez Alonso, Andr\'{e} Martins, Jan Ma\v{s}ek, Yuji
  Matsumoto, Ryan {McDonald}, Anna Missil\"{a}, Verginica Mititelu, Yusuke
  Miyao, Simonetta Montemagni, Amir More, Shunsuke Mori, Bohdan Moskalevskyi,
  Kadri Muischnek, Nina Mustafina, Kaili M\"{u}\"{u}risep, Luong~Nguy{\~{\^e}}n
  Th{\d i}, Huy{\`{\^e}}n Nguy{\~{\^e}}n~Th{\d i} Minh, Vitaly Nikolaev, Hanna
  Nurmi, Stina Ojala, Petya Osenova, Lilja {{\O}}vrelid, Elena Pascual, Marco
  Passarotti, Cenel-Augusto Perez, Guy Perrier, Slav Petrov, Jussi Piitulainen,
  Barbara Plank, Martin Popel, Lauma Pretkalni\c{n}a, Prokopis Prokopidis,
  Tiina Puolakainen, Sampo Pyysalo, Alexandre Rademaker, Loganathan Ramasamy,
  Livy Real, Laura Rituma, Rudolf Rosa, Shadi Saleh, Manuela Sanguinetti, Baiba
  Saul{\={\i}}te, Sebastian Schuster, Djam\'{e} Seddah, Wolfgang Seeker, Mojgan
  Seraji, Lena Shakurova, Mo~Shen, Dmitry Sichinava, Natalia Silveira, Maria
  Simi, Radu Simionescu, Katalin Simk\'{o}, M\'{a}ria \v{S}imkov\'{a}, Kiril
  Simov, Aaron Smith, Alane Suhr, Umut Sulubacak, Zsolt Sz\'{a}nt\'{o}, Dima
  Taji, Takaaki Tanaka, Reut Tsarfaty, Francis Tyers, Sumire Uematsu, Larraitz
  Uria, Gertjan van Noord, Viktor Varga, Veronika Vincze, Jonathan~North
  Washington, Zden\v{e}k \v{Z}abokrtsk\'{y}, Amir Zeldes, Daniel Zeman, and
  Hanzhi Zhu.
\newblock 2017.
\newblock Universal dependencies 2.0.
\newblock Technical report, {LINDAT}/{CLARIN} digital library at the Institute
  of Formal and Applied Linguistics (\'{U}FAL), Faculty of Mathematics and
  Physics, Charles University.

\bibitem[\protect\citename{Popel \bgroup et al.\egroup
  }2017]{PopelZabokrtskyVojtek2017}
Martin Popel, Zdenek Zabokrtsk\'{y}, and Martin Vojtek.
\newblock 2017.
\newblock Udapi: Universal {API} for {U}niversal {D}ependencies.
\newblock In {\em Universal Dependencies Workshop at NoDaLiDa 2017},
  Gothenburg.

\bibitem[\protect\citename{Santorini}1990]{Santorini1990}
Beatrice Santorini.
\newblock 1990.
\newblock Part-of-speech tagging guidelines for the {Penn Treebank} project
  (3rd revision).
\newblock Technical report, University of Pennsylvania, University of
  Pennsylvania.

\bibitem[\protect\citename{Schuster and Manning}2016]{SchusterManning2016}
Sebastian Schuster and Christopher~D. Manning.
\newblock 2016.
\newblock Enhanced {E}nglish {U}niversal {D}ependencies: An improved
  representation for natural language understanding tasks.
\newblock In {\em Proceedings of LREC 2016}, pages 2371--2378, Portoro\v{z},
  Slovenia.

\bibitem[\protect\citename{Schuster \bgroup et al.\egroup
  }2018]{SchusterNivreManning2018}
Sebastian Schuster, Joakim Nivre, and Christopher~D. Manning.
\newblock 2018.
\newblock Sentences with gapping: Parsing and reconstructing elided predicates.
\newblock In {\em Proceedings of NAACL 2018}, pages 1156--1168, New Orleans,
  LA.

\bibitem[\protect\citename{Silveira \bgroup et al.\egroup
  }2014]{SilveiraEtAl2014}
Natalia Silveira, Timothy Dozat, Marie-Catherine de~Marneffe, Samuel~R. Bowman,
  Miriam Connor, John Bauery, and Christopher~D. Manning.
\newblock 2014.
\newblock A gold standard dependency corpus for {E}nglish.
\newblock In {\em Proceedings of the Ninth International Conference on Language
  Resources and Evaluation (LREC-2014)}, pages 2897--2904, Reykjavik, Iceland.

\bibitem[\protect\citename{Zeldes and Simonson}2016]{ZeldesSimonson2016}
Amir Zeldes and Dan Simonson.
\newblock 2016.
\newblock Different flavors of {GUM}: Evaluating genre and sentence type
  effects on multilayer corpus annotation quality.
\newblock In {\em Proceedings of LAW X – The 10th Linguistic Annotation
  Workshop}, pages 68--78, Berlin.

\bibitem[\protect\citename{Zeldes}2017]{Zeldes2017}
Amir Zeldes.
\newblock 2017.
\newblock The {GUM} {C}orpus: Creating multilayer resources in the classroom.
\newblock {\em Language Resources and Evaluation}, 51(3):581--612.

\end{thebibliography}
\bibliographystyle{acl}
\end{document}